# Deep Mangoes: from fruit detection to cultivar identification in colour images of mango trees.

Philippe Borianne
CIRAD, UMR AMAP, F-34398 Montpellier, France
AMAP, Univ Montpellier, CIRAD, CNRS, INRA, IRD, Montpellier, France
philippe.borianne@cirad.fr

Julien Sarron
CIRAD, UPR HortSys, F-34398 Montpellier, France
Centre pour le Développement de l'Horticulture, ISRA, Dakar, Senegal
Univ Montpellier, F-34090 Montpellier, France
julien.sarron@cirad.fr

Frédéric Borne
CIRAD, UMR AMAP, F-34398 Montpellier, France
AMAP, Univ Montpellier, CIRAD, CNRS, INRA, IRD, Montpellier, France
frederic.borne@cirad.fr

Émile Faye
CIRAD, UPR HortSys, F-34398 Montpellier, France
Centre pour le Développement de l'Horticulture, ISRA, Dakar, Senegal
Univ Montpellier, F-34090 Montpellier, France
emile.faye@cirad.fr

## ABSTRACT

This paper presents results on the detection and identification mango fruits from colour images of trees. We evaluate the behaviour and the performances of the Faster R-CNN network to determine whether it is robust enough to "detect and classify" fruits under particularly heterogeneous conditions in terms of plant cultivars, plantation scheme, and visual information acquisition contexts. The network is trained to distinguish the 'Kent', 'Keitt', and 'Boucodiekhal' mango cultivars from 3,000 representative labelled fruit annotations. The validation set composed of about 7,000 annotations was then tested with a confidence threshold of 0.7 and a Non-Maximal-Suppression threshold of 0.25. With a F1-score of 0.90, the Faster R-CNN is well suitable to the simple fruit detection in tiles of 500x500 pixels. We then combine a multi-tiling approach with a Jaccard matrix to merge the different parts of objects detected several times, and thus report the detections made at the tile scale to the native 6,000x4,000 pixel size images. Nonetheless with a F1-score of 0.56, the cultivar identification Faster R-CNN network presents some limitations for simultaneously detecting the mango fruits and identifying their respective cultivars. Despite the proven errors in fruit detection, the cultivar identification rates of the detected mango fruits are in the order of 80%. The ideal solution could combine a Mask R-CNN for the image pre-segmentation of trees and a double-stream Faster R-CNN for detecting the mango fruits and identifying their respective cultivar to provide predictions more relevant to users' expectations.

## 1. INTRODUCTION

The estimation of pre-harvest agricultural production is essential to meet development challenges and reduce the vulnerability of populations to global changes. Indeed, one of the main issues hindering the development of perennial crops is the impossibility of early, easily, and accurately estimating crop yields in order to guide farm management effectively. To date, yield estimation in tropical orchards is still based on a visual inspection of a limited sample of trees, a tedious, time- and cost-consuming method that depends on the reliability and accuracy of the observer [1]. The cultivation of mango (*Mangifera indica* L.) in West Africa, and more particularly in Senegal, needs particular attention because of the physiological specificities of trees, especially the reproductive asynchronism and the inter-tree heterogeneity [2]. Moreover, West African farmers require tools adapted to the conditions of their family-sized and diversified cropping systems that can provide them with production information in a simple, inexpensive and convenient way.

Over the last few years, several studies described efficient machine vision systems for fruit detection and fruit yield estimation [3]. These systems ranged from simple colour pixel segmentation to more advanced machine learning method using combination of colour, shape and texture features acquired by multiple sensors. Moreover, they were usually developed and evaluated in orchards under homogeneous conditions in terms of plant cultivars, plantation schemes (*density, row, etc.*), cultivation practices and visual information acquisition methods [4, 5, 6]. Under heterogeneous conditions, the performance of machine learning approaches drops significantly [7].

One of the major advances in the object localization and detection was the multi-scale sliding window algorithm [8, 9] for using Convolutional Neural Networks (CNNs). Regions with CNN features or R-CNN [10] improved by almost 50% the detection performance by combining the object extraction using a Selective Search [11] and the region classification by SVMs [12]. This key step was published under the name Fast R-CNN [13]. As the R-CNN, it used Selective Search to extract possible objects, but instead of typing them using SVM classifiers, it applied the CNN on all the image and then used both Region of Interest (RoI) and Pooling on the feature map with a final feed forward network for classification and regression. The RoI Pooling layer and the fully connected layers allowed the model to be end-to-end differentiable and easier to train. Faster R-CNN [14, 15] added a Region Proposal Network (RPN), in an attempt to do without the Selective Search algorithm and make the model completely trainable end-to-end.

Different works used deep CNN for yield estimation [16] and/or fruit counting [17], and especially based on the Faster R-CNN [18, 19]. Despite the asynchronous production cycles of mango trees, this neural network obtained high performances for fruit detection and counting at plot scale in homogeneous field conditions [20, 21].

To our knowledge, no work has been done to identify the stage of development of the fruits and their cultivar, which is assumed to be known. However, these two aspects are essential for estimating economic yields and forecasts of fruit farms, particularly on family farms that are increasingly using multiple grafts: several fruit cultivars are simultaneously grafted on the same tree. One of the major interests of this type of practice is to be able to diversify the production on a small area in order to make the farmer less dependent on the fluctuations of economic markets while preserving biodiversity and the health balance of the crops.

Besides mango detection in heterogeneous mango orchards, this study presents the results of fruits classification among mango cultivars. Indeed, visual images of mango trees often encompassed fruits carried by other nearby trees (adjacent or in background), and fruits too heavily occluded so that their status is difficult to assess. In order to provide farmers with the most accurate results of the number of fruits that will actually be produced per tree, we investigated the capacities of a Faster R-CNN network to classify the fruit condition.

The evaluation of the fruit detection and identification is usually done using the standard Mean Average Precision [22]. Among these indicators, we used the F1-score well suited to the statistical comparison with a real "image truth". Different notions were introduced, for example the global Jaccard indicator for qualifying the geometric fitting between expert annotations and network predictions or the multi-tiling for aggregating multiple detections. The detection reporting from tiles to images is also described and discussed.

## 2. MODEL AND METHODS
### 2.1 The Faster R- CNN

Faster R-CNN started with 500x500 pixel pre-cut images called tiles and provided a list of labelled bounding boxes with prediction probabilities. Its architecture is based on three components (see Figure 1).

First, a Region Proposal Network (RPN) is used to find up to a predefined number of regions that may contain objects. Fixed sized reference bounding boxes are placed uniformly throughout the original image. They give a list of possible relevant objects and their locations in the original image.

Second, a Region of Interest Pooling (RoIP) is applied using the features extracted by the CNN in the previously given bounding boxes to extract those features which would correspond to the relevant objects. This step significantly accelerates the identification of the interest areas.

Third, the R-CNN module uses that information to classify the content in the bounding box (or discard it using the "background" label) and adjust the bounding box coordinates (so it better fits the object).

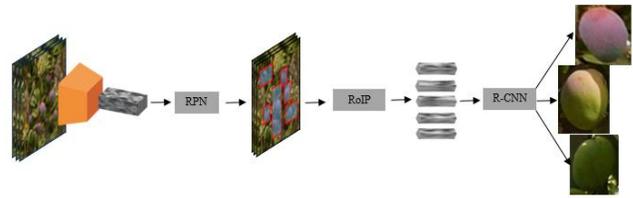

**Figure 1. Complete Faster R-CNN architecture.**

This method involves using a CNN pre-trained for the task of classification and then re-trained for the recognition of mangoes. No real consensus exists on the best network architecture for the pre-training step, i.e. between different networks with a varying number of weights. Even if the ZF [23], MobileNet [24] and DenseNet [25] pre-training seem attractive, we used the VGG [26] pre-trained on ImageNet [27] because it was dedicated to the large-scale image recognition. The network has been fine-tuned from a learning set of annotated tiles, where the annotations are bounding boxes surrounding visible (part of) mango fruits and labelled according to the defined classes.

### 2.2 Key network parameters

**The confidence score.** The network gives each box the probability of containing an object. The confidence-threshold is set to detect only boxes with high certainty.

#### 2.2.1 The non-maximum suppression [28]
The Non-Maximum Suppression (NMS) pre-vents multiple detection of the same object. The NMS-threshold is set to define the non-significant intersections of boxes.

#### 2.2.2 The iteration number
The main mechanism of networks, called gradient descent, is based on an iterative minimum search algorithm: the algorithm needs to be repeated to converge to the best possible solution. The number of iterations (given as a train-ing parameter) indirectly defines the number of times the network observes the data.

### 2.3 Matching Function

Matching function consists in associating two by two the expert annotation and network prediction boxes. The optimum matching between the expert annotations $A_i$ and the network predictions $P_j$ was given by the Jaccard matrix [29]; the $J_{i,j}$ Jaccard index was defined as the ratio of areas between both the intersection and the reunion of two boxes (1). Matches were made in descending order of Jaccard indices as long as greater than 0.25. Once fitted, the boxes were removed from the list of possible candidates and the process was repeated until all the boxes have been matched or no box could be matched.

$$J_{i,j} = \frac{area(A_i \cap P_j)}{area(A_i \cup P_j)} \quad (1)$$

The algorithmic complexity of the matching was in O(IJ) where I and J respectively represented the maximum numbers of annotations and predictions per tile (which visibly does not exceed 20 objects).

### 2.4 Merging Function

Image tiling, i.e. partitioning the image into tiles which can be processed by the neuronal network, is necessary for the detection of "small" objects in "large" images. Without tiling, small objects "disappear" during the geometric transformations applied for data formatting. If the recomposition of the image is simply achieved by juxtaposing the tiles, the aggregation of detected objects is potentially more complex, especially for objects "cut" into several parts by the tiling. When sufficiently significant i.e. quite close to the examples given in the training data set, these different parts are respectively detected, and the object is counted several times (see Figure 2). When not significant enough, no part of the object is detected and the object is not recorded. The multiple detections of the same object appearing on several tiles must be merged to adjust as well as possible the count and the shape of the boxes characterizing this last one.

We proposed an approach based on the following assumption: "The size of the objects to be detected is significantly smaller than those of the tiles used. Consequently, there should be at least one tile in which the object appears integrally". We used so different tiling, i.e. ways to partition the native image. For a given tiling, the objects validated as detected (or detections) were those whose bounding box was not incident at any edges of the tile that contained it. Objects not checking this condition were not retained: they were detected in another tiling, more exactly in a more appropriate tile containing them integrally. The problem was thus to merge detections coming from different tiling while taking care not to count more than once a given object.

The matching function previously described, based on the use of a Jaccard matrix, was used here to identify matches between objects detected in two different tiling. In case of matching, the smaller bounding box was removed. Objects that find no homologue correspond to the identification of "integral" objects in one of the tiling and "cut" objects in the other.

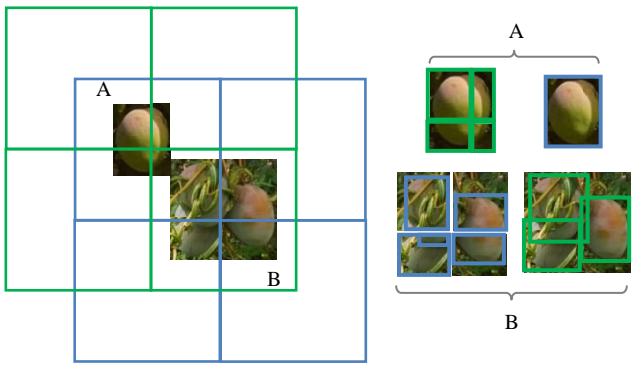

**Figure 2. Image tiling and object detection. A- The green tiling splits the object into 4 parts which are each detected by the network; in the blue tiling, the object is detected only once in its entirety. B- The blue tiling breaks 3 objects into 4 parts leading to 5 detections; in the green tiling, the 3 objects are correctly detected.**

## 2.5  Merging Function

### 2.5.1  The statistical indicator

We used the F1-score (2) as test's accuracy between 'image truth' and network predictions; it is defined as the best compromise between precision (3) and recall (4), two statistical indicators taking into account respectively the numbers of "True Positive", "False Positive" and "False Negative" classes where

- "True Positives" (TP) are mangoes annotated by the expert and correctly detected by the network,
- "False Negatives" (FN) are mangoes annotated by the expert and not detected by the network,
- "False Positives" (FP) are mangoes detected by the network but not considered as such by the expert.

$$F1score = \frac{2 \times precision \times recall}{precision + recall} \quad (2)$$

$$precision = \frac{number\ of\ mangoes\ correctly\ detected}{total\ number\ of\ detected\ mangoes} = \frac{VP}{VP+FP} \quad (3)$$

$$recall = \frac{number\ of\ mangoes\ correctly\ detected}{total\ number\ of\ epert\ anntated\ mangoes} = \frac{VP}{VP+FN} \quad (4)$$

### 2.5.2  The geometric fitting

A global Jaccard index (5) was introduced to for quantifying the difference between the geometry of the $A_i$ expert annotations and the $P_i$ network predictions, i.e. the precision of the geometric fitting of the bounding boxes of mangoes.

$$J_\alpha = \frac{\sum_i area(A_i \cap P_i)}{\sum_i area(A_i \cup P_i) + \alpha \sum_j area(A_j) + \alpha \sum_k area(P_k)} \quad (5)$$
$$where\ i \in \{VP\}, j \in \{FN\}, k \in \{FP\}$$

An index of 1 will be the perfect superposition of annotations and prediction boxes.

## 3.  RESULTS AND DISCUSSION

The main question is whether the network is sensitive enough to identifying the cultivar of the mango fruits it detects.

We used 150 native colour images of *'Kent', 'Keitt', and 'Boucodiekhal' (Bdh)* mango cultivars taken at a distance of 5 meters by a Sony Nex-7 RGB camera with a fixed focal length of 18 mm. The 4000x6000 pixel native images were cut in 500x500 pixel tiles which were manually annotated and labelled under ImageJ [30] to form a representative set of about 10,000 annotations in terms of shape, colour, sunlight conditions or occlusions. The labelled annotations were rectangular boxes, from 10 to 80 pixels side, including as closely as possible the visible fruits on the images: labels specified mango cultivar. These annotated data have been distributed as follows: 3,000 heterogeneous annotations for network training, 7,000 for cross validation.

### 3.1  Parameter setting

Various experiments were conducted to determine the best settings, i.e. the ones that maximizes the performances of the network in detection mango fruit. The same set-tings were used for fruit cultivar identification.

### 3.1.1  The confidence and NMS levels

The fruit detection network was trained with an arbitrarily number of 50,000 iterations. The validation set was then tested by simultaneously varying the confidence and NMS thresholds from 0.35 to 0.9 and 0.005 to 0.5 respectively in steps of 0.05 (see Figure 3). The F1-score (2) ranged from 0.66 for the threshold-couple (0.35, 0.5) to 0.90 – in fact 0.899687 – for (0.7, 0.25).

Different settings led to relatively close performances. We chose the setting resulting in the highest performances, acknowledging that this result certainly depended on both the training and the validation sets used. Experimentation could have been further developed by studying the performance of the cultivar identification network to find the optimum threshold pair values, but we considered that the optimum parameter setting of the fruit detection network could be transposed to the cultivar identification network.

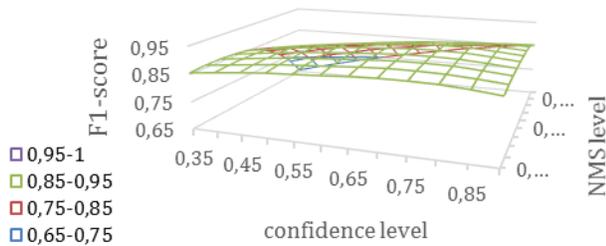

**Figure 3. Variation in the network performance according to the confidence and NMS thresholds.**

### 3.1.2 *The number of iteration*
The network was trained with numbers of iterations ranging from 500 to 30,000 in steps of 500. The validation set was then tested with a confidence threshold of 0.7 and a NMS threshold of 0.25 (see Figure 4).

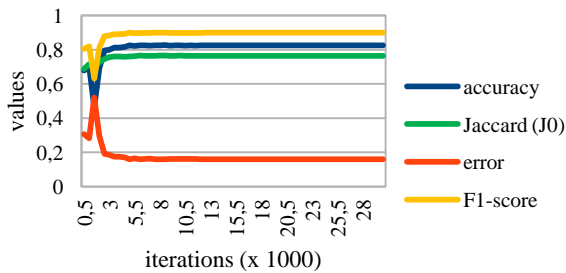

**Figure 4. Variation in the network performances according to the number of iterations.**

Error and accuracy were indicative: they represented respectively the percentage of 'correct' and 'incorrect' predictions by the network. The functions characterizing the network behaviour were strongly correlated and were stabilized from 15,000 iterations.

### 3.1.3 *The size of the training data set*
The network was trained with an iteration number of 15,000 and a size of the training data set ranging from 200 to 5,000 by steps of 200. The validation set was then tested with a confidence threshold of 0.7 and a NMS threshold of 0.25 (see Figure 5).

The network performance reached its maximum after 2,650 training annotations: the accuracy of the geometric fitting did not exceed 77%, even when significantly increasing the size of the training data set. A training set of 3,000 annotations reinforced the stability of the detection process.

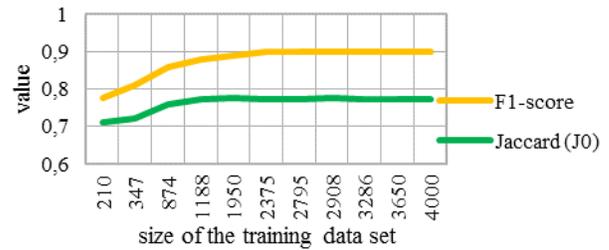

**Figure 5. Variation in the network performances according to the size of the training data set.**

### 3.2 Localization and identification of fruits
In the orchards of Western Africa, fruit plots often mix different cultivars of fruit for a better use of soils and to spread the fruit production along the year. Estimating mango yield at the orchard scale requires assessing the production of each cultivar grown, i.e. the number of fruits produced (or to be produced) per cultivar and tree.

For all the experiments below, the network was fine-tuned with 3,000 annotations and 15,000 iterations. The validation set was then tested with a confidence threshold of 0.7 and a NMS threshold of 0.25.

### 3.2.1 *The mango detection at the tile scale*

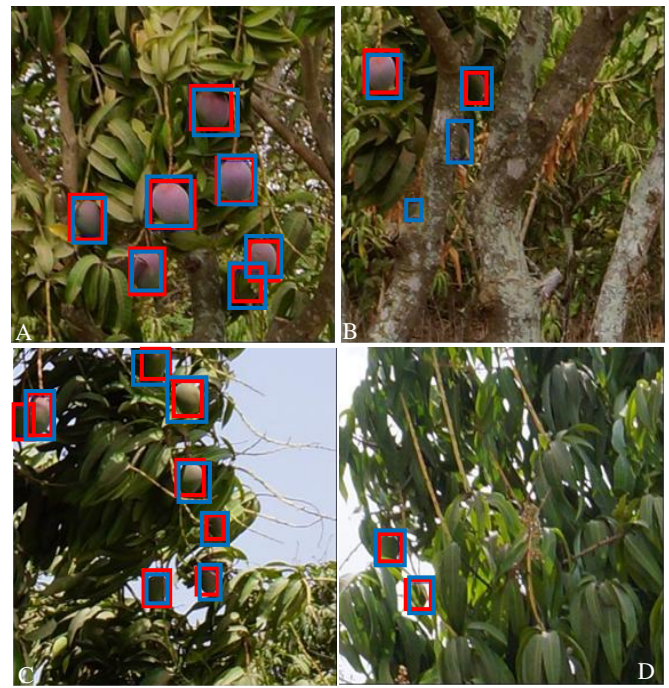

**Figure 6. Detection of mangoes in tree tiles. Red boxes were expert annotated fruits, and blue boxes network detected fruits. The network detected visible mangoes well, regardless of light exposure or fruit growing stage. The pictures shown tiles of mango trees seen at 5 meters distance for 'Keitt' (A, B) and 'Kent' (C, D) cultivars.**

This experiment aimed to ensure that the network can satisfactorily locate mangoes of various cultivars and stages of development in images of full trees.

The F1-score on the validation set was 0.90 for a J0 fitting precision of 0.77 (see Figure 4): the geometric offset between annotations and predictions could appear high, but it was a very satisfactory result considering the heterogeneity of the data set.

In Figure 6, the expert annotations (in red) were almost superimposed on the network predictions (in blue).

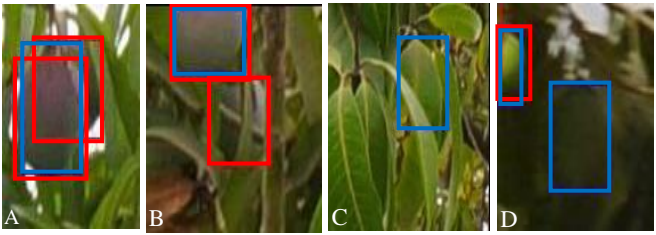

**Figure 7. Detection errors. False Negatives are red boxes, False Positives blue boxes. A- a background mango was not detected due to the application of the NMS threshold; B- mango partially hidden by foliage was not detected because this situation is not sufficiently represented in the training data set. C- a leaf was confused with a mango by the network. D- a mango inside the tree was detected by the network even though it had not been annotated by expert.**

The in-depth study of False Negatives pointed out the network's difficulties in detecting partially occulted fruits, essentially in fruit clusters (see Figure 7.A) and " fruits inside canopy (see Figure 7.B). In the first case, the NMS threshold prevented 'correct' detection of background fruits partially masked by foreground fruits. In the second case, fruits were rather hidden by leaves and wood, causing a significant change in the signature of the mango. On the False Positive side, most of the residual confusion concerned leaf arrangements with curvatures similar to those of mangoes (see Figure 7.C). Interestingly, the network detected fruits not annotated by the expert (see Figure 6.B and 7.D): in absolute terms, they were false False-Positives which penalized the network performance.

### 3.2.2 The mango detection at the full image scale

Four tiling were used to cover a maxi-mum of cutting configurations, without however being able to guarantee that they are all well considered: the 1st is a partition in tiles of 500x500 pixels starting from the image point coordinates (0,0), the second from the point (0,250), the third from the point (250,0) and the fourth from the point (250,250).

The F1-score was 0.71 for a $J_0$ fitting precision of 0.67. This drop in performance is explained by the significant increase in False Positives (see Figure 8) compared to the previous experiment: the expert limited his count to the foreground tree while the network processed all the trees in the image. Previously, only tiles in which the expert had observed a fruit were used, which partially excluded fruits from adjacent trees; here, all tiles in the images were used.

The algorithmic complexity of the detection merging is in $O(n^2)$ where n represents the number of fruits visible in the image: it is fixed by the calculation of the Jaccard matrix. It can look high even if it only takes 30" to process a native image on a Dell Studio XPS 8 x i7 3 GHz with a Nvidia Quadro M2000 graphics card. Further study to determine the optimal tiles to use for the fusion of detected fruits would be of definite theoretical interest, even if the best way to reduce computation times would be to use adaptive tiles, i.e. tiles centred on "clipped" objects: but such an approach would require two instantiations of the neural network. The major problem is to reduce the exploration area of the network to the foreground tree. The YOLO network [31] is currently under study to pre-zone the image and exclude background trees: to refine the results, it will probably be necessary to use (in addition or not) segmentation networks such as SegNet [32] or similar [33]. For example, Mask-RCNN [34] customizes the Faster-RCNN with an additional parallel branch that makes it powerful for both the bounding-box object detection and the object segmentation.

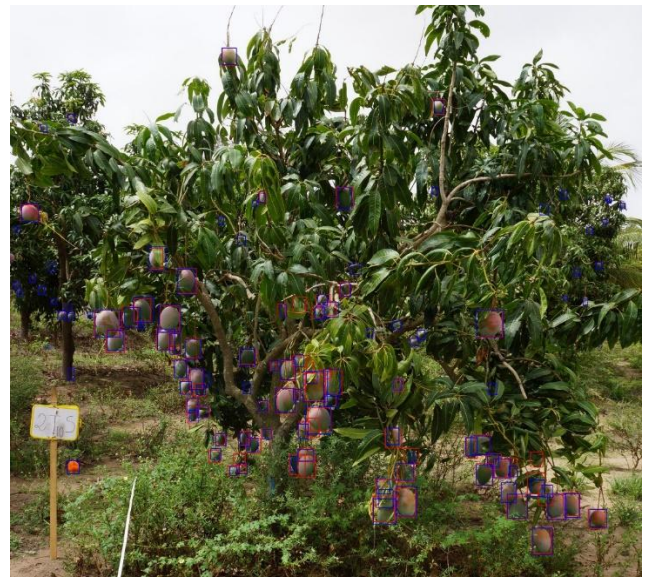

**Figure 8. Detection of mangoes in 'Keitt' tree image. Red boxes were expert annotated fruits, blue boxes network detected fruits. Unlike the network, the expert did not consider fruits of the background trees**.

### 3.2.3 Identification of mango cultivars

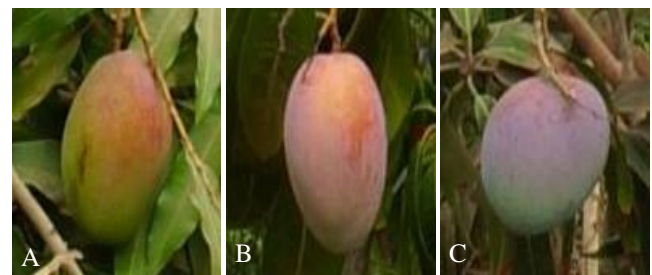

**Figure 9. Mango cultivars at the pre-harvest stage. A- 'Bdh' mango fruit is ovoid-flat with a skin col-our ranging from red to green and an average weight of 450 grams. B- 'Keitt' mango fruit is ovoid-oblong; skin color is pink with less than 30% red and the average weight is 500 to 600 grams. C- 'Kent' mango fruit is rather ovoid-wide with yellow skin and red spot; the average weight does not exceed 550 grams.**

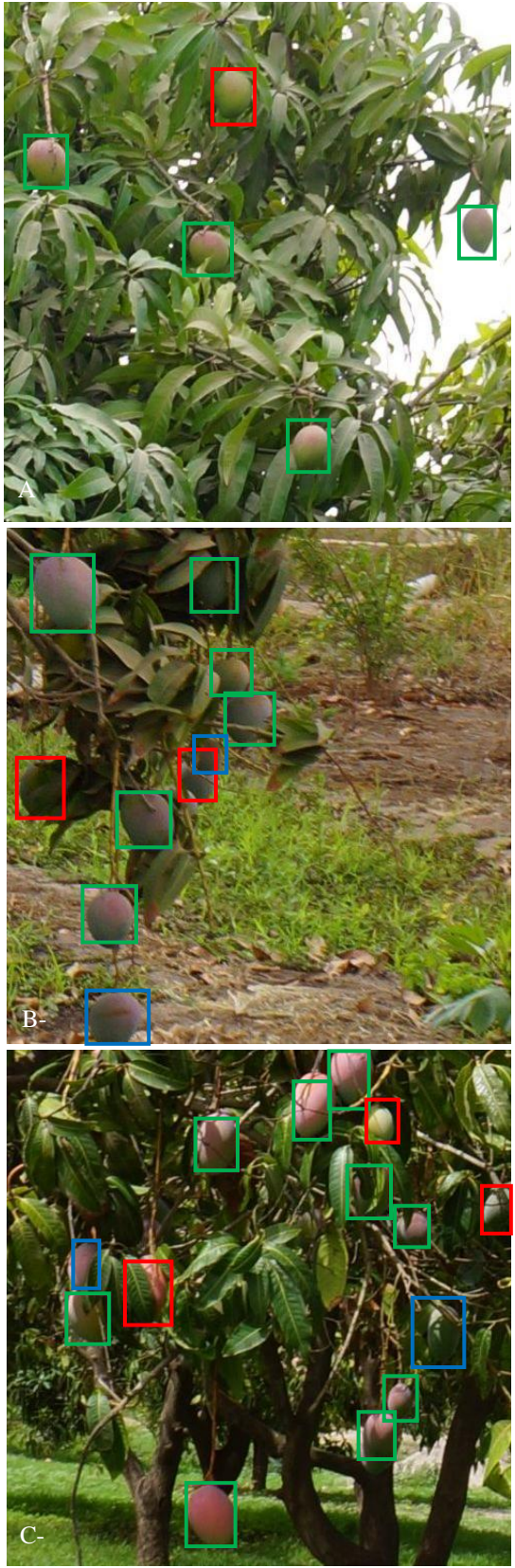

**Figure 10. Identification errors of Bdh (A), Keitt (B) and Kent (C) cultivars.**

For this experiment, the network was trained to identify the 'Kent', 'Keitt' and 'Bdh' cultivars composing the data set (see Figure 9). For the learning step, each cultivar class was described by about 1,000 labelled annotations.

In identifying cultivars, the F1-score was to 0.56 for a $J_0$ fitting precision of 0.71. This significant drop in performances is explained by the high increase in False Negatives, i.e. mangoes not detected: barely 4,000 mangoes were detected by the network during the cross-validation while the data set contains nearly 7,000 expert annotations.

When the tested fruit looks alike fruits of at least two variety classes, the network assigns predictive probabilities (by class) to it that are quite close and finally lower than the confidence threshold used by the network for the detection phase. Consequently, the object was not considered as a fruit and is therefore not detected. (see Figure 10, blue rectangles). Lowering the confidence threshold may in some cases be an option, but it might significantly increase the number of False Positives. The best option would probably be to use a two-stream faster R-CNN [35] to firstly detected all the fruits of the image and then qualify them in classes.

Figure 10 shown the behaviour of the identification network: mangoes correctly detected and identified by the cultivar identification network were represented by green rectangles, mangoes correctly detected but incorrectly identified by red rectangles, and mangoes not detected by the identification network although counted by a simple detection network by blue rectangles.

Errors in identification probably have several causes. Among them, the physiological features are not negligible. Since the expression of the really discriminating features of fruits only appears after the juvenile stages, there would therefore be a link between the cultivar identification certainty and the fruit development stages. It is therefore necessary to know / be able to identify the stage of development of the fruits to certify the detected fruit cultivars. This is all the more important in the case of grafted trees that carry several fruit cultivars. Moreover, angles of view or partial occlusions of fruits probably have a greater impact on the responses of an identification network than on those of a detection network.

Beyond this observation, the question was how many correctly detected mangoes were well identified (classified). Table 1 gave the identification percentages, class by class, of correct detected fruits (900 'Bdh' mangoes detected on 2,000 annotated, 1,200 'Keitt' on 2,500 and 1,600 'Kent' on 2,500). The identification rates without error; i.e. the detected fruits attributed to the right cultivar class, ranged from 80 to 90% (see bold values of the Table 1). The identification errors were caused by visual similarities in mangoes, especially when partially masked, viewed from irrelevant angles or at too early stages of development.

**Table 1. Multi-class identification rate.** *Each column indicates how the neural network identifies the fruits of a given cultivar.*

|  |  | Expert labelling | | |
|---|---|---|---|---|
|  |  | **Bdh** | **Keitt** | **Kent** |
| Network labelling | **Bdh** | **86.9** | 2.4 | 1.0 |
|  | **Keitt** | 2.6 | **78.2** | 6.6 |
|  | **Kent** | 10.5 | 19.4 | **92.4** |

Some questions remain open: for example, why is there 3 times more confusion between the Keitt and Kent fruits than between Kent and Keitt fruits: a simple conjuncture or a deeper network learning problem? It is obvious that, unlike adult or mature fruits, young fruits have very similar aspects, perhaps too similar, so the network can unambiguously identified their respective cultivar. But if that were possible, the solution would surely based on a Ensemble Neuronal Network [36, 37] i.e. on the simultaneous use of several different networks fine-tuned on a same dataset or on several identical networks trained on different data for significantly increasing the relevance of fruit cultivar identification unless this qualification is limited to sufficiently mature fruit.

## 4. CONCLUSION

We evaluated the behaviour of the Faster R-CNN network to determine whether it was robust enough to "detect and identify" fruits under particularly heterogeneous conditions in terms of tree cultivars, plantation scheme, and visual information acquisition contexts.

The network was trained using 3,000 representative labelled annotations of *'Kent', 'Keitt', and 'Boucodiekhal' mango cultivars*. The validation set composed of about 7,000 labelled annotations was tested with a confidence threshold of 0.7 and a Non-Maximal-Suppression threshold of 0.25.

The network accuracy (F1-score) was 90% for fruit detection, but fell to 56% for fruit cultivar identification. When the tested fruit looks alike fruits of at least two variety classes, the network assigns predictive probabilities (by class) to it that are quite close and finally lower than the confidence threshold used by the network for the detection phase. Consequently, the object was not considered as a fruit and is therefore not detected.

The cultivar identification rates of the detected mango fruits were in the order of 80%. Errors were caused by visual similarities in mangoes, especially when partially masked, viewed from irrelevant angles or at too early stages of development.

Image tiling, i.e. partitioning the image into tiles which can be processed by the neuronal network, is necessary for the detection of "small" objects in "large" images. Tiling may lead to "artificially" clipped objects, which will result in multiple detection of these objects. We combined a multi-tiling approach with a Jaccard matrix to identify and merge multiple detections and thus report the detections made at the tile scale to the native image.

The cross validations showed the need to undertake additional reflexion to make the predictions of the R-CNN Faster network more relevant to users' expectations. Future works could focused on (i) the image pre-processing to separate neighbouring trees from each other (using convolutional segmentation networks such as SegNet or Mask R-CNN) and, (ii) the two-stream Faster R-CNN developing to identify the cultivar of all detected fruits on each segmented tree.

## 5. ACKNOWLEDGMENTS

This work was supported by the French National Research Agency under the Invest-ments for the Future Program (ANR-16-CONV-0004) and by the PixYield Creativity and Scientific Innovation action funded by CIRAD, the French agricultural research and international cooperation organization. The authors wish to thank Dr. Sané C.A.B for its expert work and active participation in the annotation of visual data.